%% file: colm2024_conference.tex
\documentclass{article} 
\usepackage{colm2024_conference}

\usepackage{microtype}
\usepackage{hyperref}
\usepackage{url}
\usepackage{booktabs}

\usepackage{xspace}
\usepackage{multirow}
\usepackage{booktabs}
\usepackage{bm}
\usepackage{amsmath}
\usepackage{amsfonts}
\usepackage{hyperref}
\usepackage{graphicx}
\usepackage{framed}
\usepackage{enumitem}
\usepackage{pifont}
\usepackage{CJKutf8}
\usepackage{marvosym}
\usepackage{comment}
\usepackage{algorithm}
\usepackage{algorithmic}
\usepackage{amssymb}
\usepackage{seqsplit}
\usepackage{float}

\newcommand{\ie}{\emph{i.e.,}\xspace}

\newcommand{\eg}{\emph{e.g.,}\xspace}

\newcommand{\etc}{\emph{etc}}
\newcommand{\ignore}[1]{}

\title{Towards Coarse-to-Fine Evaluation of Inference Efficiency for Large Language Models}

\author{Yushuo Chen$^{1}$, Tianyi Tang$^{1}$\thanks{Authors contributed equally.}, Erge Xiang$^{2*}$, Linjiang Li$^{2*}$,\\
\textbf{Wayne Xin Zhao}$^{1}$\thanks{Corresponding authors.}, \textbf{Jing Wang}$^{2\dagger}$, \textbf{Yunpeng Chai}$^{2\dagger}$, \textbf{Ji-Rong Wen}$^{1,2}$ \\
$^{1}$Gaoling School of Artificial Intelligence, Renmin University of China \\
$^{2}$School of Information, Renmin University of China\\
\texttt{chenyushuo1999@foxmail.com,steventianyitang@outlook.com,}\\
\texttt{\{ergeda,lilinjiang,jwang,ypchai,jrwen\}@ruc.edu.cn,batmanfly@gmail.com}\\
}

\begin{document}

\maketitle

\begin{abstract}
In real world, large language models~(LLMs) can serve as the assistant to help users accomplish their jobs, and also support the development of advanced applications.
For the wide application of LLMs, the inference efficiency is an essential concern, which has been widely studied in existing work, and numerous optimization algorithms and code libraries have been proposed to improve it.
Nonetheless, users still find it challenging to compare the effectiveness of all the above methods and understand the underlying mechanisms.
In this work, we perform a detailed coarse-to-fine analysis of the inference performance of various code libraries.
To evaluate the overall effectiveness, we examine four usage scenarios within two practical applications. 
We further provide both theoretical and empirical fine-grained analyses of each module in the Transformer architecture.
Our experiments yield comprehensive results that are invaluable for researchers to evaluate code libraries and improve inference strategies.





\end{abstract}

\begin{CJK*}{UTF8}{gbsn}
\input{sec-intro}

\input{sec-pre}

\input{sec-overall}

\input{sec-analysis}

\input{sec-rel}

\input{sec-con}
\end{CJK*}

\bibliography{colm2024_conference}
\bibliographystyle{colm2024_conference}
\input{sec-app}

\end{document}

%% file: sec-intro.tex
\section{Introduction} 

With the advancement and wide spread of large language models~(LLMs)~\citep{LLMSurvey}, the enhancement of inference efficiency in LLMs has emerged as an important topic of contemporary research~\citep{infer-survey,infer-survey-2}. To achieve superior inference speed without significant performance degradation, researchers have proposed diverse inference optimization algorithms and libraries. 

Currently, several prominent libraries have been widely used in the market, such as vLLM~\citep{vllm}, DeepSpeed-MII~\citep{deepspeed-mii}, and TensorRT-LLM~\citep{trt-llm}, \etc. These libraries have notably elevated inference efficiency through sophisticated methodologies such as optimization algorithms and parallel computing. Nevertheless, a notable deficiency exists in the absence of a standardized evaluation benchmark for comprehensively comparing the performance across existing libraries. To address it, this work meticulously devises a series of evaluation experiments with the goal of impartially and objectively assessing the inference efficiency of each library.


Concretely, this paper clearly defines two types of evaluation experiments: coarse-grained and fine-grained. In the coarse-grained evaluation, four text generation datasets with diverse length distributions are designed to simulate various generation tasks. We then explore two practical applications: offline batch inference and network service provisioning. The former involves assessments conducted in batch mode offline while the latter pertains to real-time online service scenarios. We assess the efficiency of each library in offline inference and also evaluate their performance at different request frequencies.

In the fine-grained analysis experiment, we provide an intricate examination of the requisite number of floating-point and memory operations for each module, to acquire a more holistic comprehension of the distribution of inference time. Besides, to pinpoint the efficiency bottleneck more accurately, we introduce the concept of computational strength and conducted an in-depth efficiency performance analysis of each module based on this concept. Furthermore, to validate the theoretical analysis, two representative libraries are selected for detailed and specific time analysis testing.

In conclusion, this investigation endeavors to delve into the inference efficiency of large language models through comprehensive and objective evaluation experiments.
First, we propose a comprehensive benchmark which covers different task scenarios, and use them to evaluate different libraries in different usage scenarios, filling the gap in the inference benchmark.
Second, we propose a fine-grained complexity analysis formula for each module of LLaMA, which reflects the bottleneck in decoding by calculating FLOPs, MOPs, and arithmetic intensity, and provides direction for subsequent decoding evaluation.
Finally, we have open-source the above dataset, code, and evaluation scripts, which are available in \url{https://github.com/RUCAIBox/Coarse-to-Fine-Evaluation-of-Inference-Efficiency}.
It is anticipated that the findings of this study will not only offer valuable insights for enhancing existing inference libraries but also establish a robust groundwork for the advancement of future inference algorithms and libraries.





\begin{table*}[t]
\small
\centering
\begin{tabular}{lccccc}
\toprule
                       & \multicolumn{2}{c}{\textbf{Evaluations}}  & \multicolumn{3}{c}{\textbf{Optimization Technologies}}         \\
\textbf{Libraries}     & \textbf{\#Real Data} & \textbf{\#Syn. Data} & \textbf{KV Cache} & \textbf{FlashAttn} & \textbf{Batching} \\ \midrule
\textbf{Transformers}  &                      &                      & Vanilla           &                    &                   \\
\textbf{vLLM}          & 3                    &                      & Blocked           & √                  & √                 \\
\textbf{DeepSpeed-MII} &                      & √                    & Blocked           &                    & √                 \\
\textbf{TGI}           &                      & √                    & Blocked           & √                  & √                 \\
\textbf{TenserRT-LLM}  & 1                    & √                    & Blocked           & √                  & √                 \\
\textbf{llama.cpp}     &                      & √                    & Sequence          &                    & √                 \\
\textbf{LightLLM}      & 1                    &                      & Token             & √                  & √                 \\
\textbf{LMDeploy}      &                      & √                    & Blocked           &                    & √                 \\
\textbf{StreamingLLM}  &                      & √                    & W-Sink            &                    &                   \\ \bottomrule
\end{tabular}
\caption{
Comparison of current open-sourced LLM inference libraries, including evaluation methods and optimization technology.
In evaluation methods part, ``\#Real Data" indicates the number of real world data scenarios. ``Syn. Data" indicates synthetic data.
``KV Cache" indicates KV cache management methods:
``Vanilla" denotes naive method, ``Blocked" denotes \emph{PagedAttention}~\citep{vllm}, 
``Token" denotes \emph{token attention} and ``W-Sink" denotes \emph{window with attention sink} method~\citep{xiao2023streamingllm}. 
``FlashAttn" indicates FlashAttention~\citep{flash-attention,flash-attention-2}. 
``Batching" indicates \emph{in-flight batching}, \emph{continuous batching} or \emph{Dynamic SplitFuse}.
}
\end{table*}



%% file: sec-pre.tex
\section{Preliminary} 

\subsection{Background of Transformer}



In contemporary LLMs, the prevailing architecture is the Transformer decoder~\citep{transformer}. Utilizing the LLaMA~\citep{llama,llama2} model as a paradigmatic illustration, its design encompasses two principal components: the multi-head attention block (MHA module) and the feed-forward network (FFN module). Both of these modules are followed by an RMS normalization~\citep{Zhang-NIPS-2019-Root} and a residual network.

The MHA module transforms the input $\bm{X}$ into three matrices $\bm{Q}, \bm{K}, \bm{V}$ through different linear transformations, calculate the multi-head attention, and aggregate the results from multiple heads using the following formulas:
\begin{align}
\bm{Q}=\bm{X}\bm{W}_Q, \bm{K}&=\bm{X}\bm{W}_K, \bm{V}=\bm{X}\bm{W}_V, \label{eq:qkv} \\
\bm{O}=\operatorname{Attention}(\bm{Q},\bm{K},\bm{V}) &= \operatorname{softmax}(\frac{\bm{Q}\bm{K}^{\intercal}}{\sqrt{d}})\bm{V}, \label{eq:attn} \\
\bm{X} &= \bm{O}\bm{W}_O, \label{eq:o}
\end{align}
where $\bm{W}_Q, \bm{W}_K, \bm{W}_V, \bm{W}_O \in \mathbb{R}^{h \times h}$ denote learnable parameters.


The FFN module uses the SwiGLU activation function~\citep{Shazeer-arxiv-2020-GLU} to expand the intermediate state dimension with gated linear units, and then obtains the output result of the module through a linear transformation:
\begin{align}
\bm{X} = [ \operatorname{Swish}(\bm{X}\bm{W}_G) \odot  (\bm{X}\bm{W}_U) ] \bm{W}_D, \label{eq:ffn}
\end{align}
where $\odot$ is Hadamard product and $\bm{W}_G, \bm{W}_U \in \mathbb{R}^{h \times h'}$ and $\bm{W}_D \in \mathbb{R}^{h' \times h}$ denote parameters.

\begin{algorithm}
\caption{Greedy search with KV cache}
\label{alg-greedy-search-kv-cache} 
\begin{algorithmic}[1]
    \REQUIRE $\textrm{Model } \mathcal{M}, \textrm{input token id list } \bm{x}$
    \ENSURE $\textrm{Response token id list } \bm{y}$
    \STATE $P, \bm{K}_{past}, \bm{V}_{past} = \mathcal{M}(\bm{x})$
    \STATE $x' = \arg \max P$
    \STATE $\bm{x} \gets \bm{x} \oplus [x']$
    \WHILE{$x'$ is not EOS $|\bm{x}|$ $\leq$ max-length} 
        \STATE $P, \bm{K}, \bm{V} = \mathcal{M}(x', \bm{K}_{past}, \bm{V}_{past})$
        \STATE $x' = \arg \max P$
        \STATE $\bm{x} \gets \bm{x} \oplus [x']$
        \STATE $\bm{K}_{past}, \bm{V}_{past} \gets \bm{K}_{past} \oplus \bm{K}, \bm{V}_{past} \oplus \bm{V}$
    \ENDWHILE
    \STATE $\bm{y} \gets \bm{x}$
\end{algorithmic} 
\end{algorithm}

After training, the inference of LLMs typically involves auto-regressive generation. 
Algorithm~\ref{alg-greedy-search-kv-cache} represents an enhancement of auto-regressive generation, delineated into two distinct phases: the \emph{prefill} phase and the \emph{decoding} phase. During the prefill phase (lines 1-3), the model generates the initial token and stores the $\bm{K}$ and $\bm{V}$ matrices corresponding to the input tokens, called \emph{KV cache}~\citep{kv-cache}.
Subsequently, in the decoding phase (lines 4-9), the model iteratively generates the next token by reusing the KV cache and updates the cache for future $\bm{K}$ and $\bm{V}$ matrices.


\subsection{Arithmetic Intensity} 
\label{arithmetic-intensity}

In model inference, temporal overhead mainly stems from GPU computation and memory access. Computation volume is measured in \emph{floating-point operations (FLOPs)}, and memory access in \emph{bytes of reads and writes (MOPs)}~\citep{infer-survey}.
Furthermore, the concept of \emph{arithmetic intensity}~\citep{GPGPU} is introduced as the ratio of the FLOPs to MOPs:
\begin{align}
\mathrm{Arithmetic\ Intensity} = \frac{\mathrm{\# FLOPs}}{\mathrm{\# MOPs}} .
\end{align}

Each computational operation (\eg linear transformation) and hardware component (\eg GPU) possesses a arithmetic intensity. When the arithmetic intensity of an operation surpasses that of the GPU, it suggests that the operation's efficiency is constrained by the GPU's computational capacity, defining a \emph{compute-bound} scenario. Conversely, if the operation's intensity is lower than the GPU's, it implies that the limitation is due to the GPU's memory bandwidth, characterizing a \emph{memory-bound} scenario.

Given this background, we are poised to undertake a comprehensive evaluation of existing inference libraries through both overall (Section~\ref{sec:overall}) and fine-grained analyses (Section~\ref{sec:fine}). This dual approach allows us to thoroughly assess the performance of LLMs decoding and identify its primary bottlenecks.

%% file: sec-overall.tex
\section{Overall Evaluation and Analysis} 
\label{sec:overall}

In this section, we conduct an overall evaluation of the inference efficiency of LLMs. We introduce a series of evaluation datasets tailored for two distinct usage scenarios.




\subsection{Evaluation Scenarios}
We examine two real-world usage scenarios: batch inference and server-based inference, across four specially constructed datasets to encompass a range of task scenarios.







\subsubsection{Task Scenarios}

We develop four datasets focusing on the generation tasks in various real-world scenarios.
The input-output length distribution of these datasets is shown in Figure~\ref{fig:length}.

$\bullet$ \textbf{Short-to-Short Dataset.}
This dataset encompasses scenarios such as question answering and daily assistance, characterized by brief inputs and outputs. We meticulously select 1,000 examples from the Alpaca dataset~\citep{alpaca}, ensuring that both the input and output lengths predominantly remain under 50 tokens.



$\bullet$ \textbf{Short-to-Long Dataset.}
Tailored for tasks like math problem solving and code generation, this dataset comprises scenarios with short inputs and more lengthy outputs. From the Alpaca dataset, we curate 1,000 instances where the input length does not exceed 50 tokens, while the output length varies up to 1,000 tokens.



$\bullet$ \textbf{Short-to-16k Dataset.}
Building on the concept of the short-to-long dataset, we delve into scenarios demanding exceptionally long-text generation, such as story generation. We select instances from the Vicuna dataset~\citep{vicuna2023}, requiring the model to produce outputs of exactly 16,000 tokens.


$\bullet$ \textbf{Long-to-Short Dataset.}
Aimed at reflecting text summarization or multi-turn dialogue scenarios, this dataset features lengthy inputs with concise outputs. Compiled from the ShareGPT dataset~\citep{sharegpt}, it includes examples where the input ranges from 1,100 to 1,500 tokens and the output is limited to 120 tokens or less.


\subsubsection{Usage Scenarios}

We mainly consider the following two usage scenarios:

$\bullet$ \textbf{Batching Inference.}
In evaluating the capabilities of LLMs, it is necessary to process extensive amounts of input data in bulk offline. This context does not require a specific order or delay to process each input, allowing for the flexible arrangement of generation sequences. We employ the four datasets to assess the time taken by different libraries to process the entire dataset, along with the token throughput.



$\bullet$ \textbf{Serving Inference.}
Contrary to batch inference, which is mainly used in research scenarios, serving inference is predominantly utilized in the network deployment to facilitate applications akin to ChatGPT. The metrics for this scenario include sequence and token throughput, measuring the system's efficacy in managing data sequences and tokens, respectively. To account for initial stabilization and concluding operations within the system, our analysis omits the first and last 100 requests. The evaluation allows for an in-depth investigation into how various libraries fare under simulated network service conditions, elucidating their capacity to manage varying loads and respond within acceptable timeframes.

\subsection{Evaluation Setup} 

$\bullet$ \textbf{Libraries.}
The libraries under evaluation encompass Transformers~(TRF)~\citep{wolftransformers}, vLLM~\citep{vllm}, Deepspeed-MII~(MII)~\citep{deepspeed-mii}, TensorRT-LLM~(TRT)~\citep{trt-llm}, and llama.cpp~(L.CPP)~\citep{llama.cpp}.
For batching inference, we manually set the batch size for TRF and employ built-in batching strategies for the other four libraries. For serving inference, we evaluate the performance of vLLM and MII.


$\bullet$ \textbf{LLMs.}
We utilize four models for evaluation: \texttt{\seqsplit{Llama-2-7b-chat-hf}}, \texttt{\seqsplit{Llama-2-13b-chat-hf}}~\citep{llama2}, \texttt{\seqsplit{vicuna-7b-v1.5-16k}}, and \texttt{\seqsplit{vicuna-13b-v1.5-16k}}~\citep{vicuna2023}. The LLaMA-2 models, which are widely used in chat applications, are chosen to assess their performance across three scenarios: short-to-short (S2S), short-to-long (S2L), and long-to-short (L2S). The Vicuna models, designed for handling long contexts, are employed to evaluate performance on the short-to-16k (S-16k) dataset. 

$\bullet$ \textbf{Hardwares.}
To assess the influence of various hardware platforms on influence efficiency, we conduct experiments using three NVIDIA GPUs: \texttt{RTX-3090}, \texttt{RTX-4090}, and \texttt{A800}. Table~\ref{tab-hardwares} presents key specifications of these GPUs, encompassing GPU memory capacity, memory bandwidth, and BF16 floating-point operations (FLOPs) per second.

\begin{table}[!t]
\small
\centering
\begin{tabular}{lccrrrrr}
\toprule
\textbf{Data}                   & \textbf{Hardware} & \textbf{Model Size} & \textbf{TRF} & \textbf{vLLM} & \textbf{MII} & \textbf{TRT} & \textbf{L.CPP} \\ \midrule
\multirow{4}{*}{\textbf{S2S}}   & \textbf{3090}     & \textbf{7B}         & 98.14        & 23.85         & 27.66        & 73.36        & 49.21          \\
                                & \textbf{4090}     & \textbf{7B}         & 70.84        & 13.89         & 27.05        & 58.79        & 83.74          \\
                                & \textbf{A800}     & \textbf{7B}         & 65.09        & 12.39         & 18.53        & 41.62        & 41.81          \\
                                & \textbf{A800}     & \textbf{13B}        & 248.46       & 24.33         & 29.98        & 76.41        & 39.70          \\ \midrule
\multirow{4}{*}{\textbf{S2L}}   & \textbf{3090}     & \textbf{7B}         & 4762.62      & 567.79        & 792.67       & 1342.81      & 1590.07        \\
                                & \textbf{4090}     & \textbf{7B}         & 5600.64      & 427.99        & 713.94       & 1206.16      & 1688.04        \\
                                & \textbf{A800}     & \textbf{7B}         & 4876.83      & 177.91        & 597.84       & 760.17       & 1271.06        \\
                                & \textbf{A800}     & \textbf{13B}        & 5879.23      & 256.03        & 825.18       & 1419.02      & 1036.68        \\ \midrule
\multirow{4}{*}{\textbf{L2S}}   & \textbf{3090}     & \textbf{7B}         & 1177.80      & 441.62        & 485.65       & 540.80       & 695.22         \\
                                & \textbf{4090}     & \textbf{7B}         & 864.07       & 269.55        & 329.86       & 294.15       & 876.12         \\
                                & \textbf{A800}     & \textbf{7B}         & 756.04       & 166.84        & 236.08       & 197.03       & 2559.78        \\
                                & \textbf{A800}     & \textbf{13B}        & 3076.05      & 369.72        & 893.91       & 360.94       & 2879.51        \\ \midrule
\multirow{2}{*}{\textbf{S-16k}} & \textbf{A800}     & \textbf{7B}         & 50566.78     & 5980.50       & 6913.22      & 10464.32     & 36158.40       \\
                                & \textbf{A800}     & \textbf{13B}        & 75257.35     & 11074.52      & 14186.84     & 33659.65     & 46040.82       \\ \bottomrule
\end{tabular}
\caption{
The total time cost in seconds for batch inference using LLaMA-2 (7B) and (13B).
}
\label{exp-batch}

\end{table}

\begin{figure*}[t]
    \centering
    \includegraphics[width=1\textwidth]{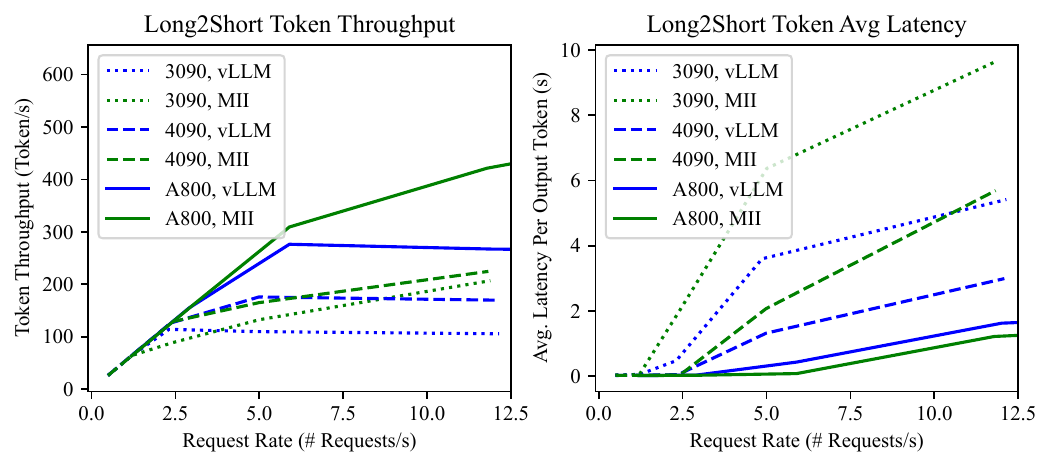}
    \caption{
        The throughput and latency for serving inference of vLLM and MII using LLaMA-2 (7B) under different request frequencies on the Long2Short dataset.
    }
    \label{fig:serving}
\end{figure*}

\subsection{Evaluation Results}




Firstly from the scenario of batching inference in Table~\ref{exp-batch} and Table~\ref{exp-batch-throughput}, we find that GPU computational performance is pivotal for short input-output pairs, whereas memory bandwidth becomes critical as sequences elongate. We have observed that the 4090 significantly outperforms the 3090 in the S2S dataset. However, this advantage diminishes in S2L and L2S datasets. Conversely, the A800 consistently excels across all datasets. 
We hypothesize that the observed performance discrepancies are attributable to the differing specifications of the GPUs (see Table~\ref{tab-hardwares}). The 4090 boasts double the computational power of the 3090, yet their bandwidths are comparable. In contrast, the A800 doubles both the computational power and bandwidth relative to the 4090.


Secondly, vLLM and MII demonstrate superior efficiency compared to other libraries in batching scenarios. This advantage is primarily attributable to their advanced optimization technologies, including KV cache and batching strategies. When analyzing the results from the 7B and 13B models, it is evident that the 13B model's processing time is nearly double that of the 7B model for both vLLM and MII. This phenomenon is not observed in other libraries. Given that the computational FLOPs for the 13B model are twice those of the 7B model, a corresponding increase in processing time is expected. This indicates that the other libraries have room for improvement in GPU memory management.

Thirdly, the Dynamic SplitFuse batching strategy of MII demonstrates enhanced efficiency in serving inference scenarios of long sequences, as evidenced by the results depicted in Figures~\ref{fig:serving},~\ref{fig:serving2}, and~\ref{fig:serving3}. It is observed that with an increasing evaluation rate of requests, the vLLM initially exhibits a surge, followed by a gradual decline after reaching its peak performance.
In contrast, the token throughput for MII consistently rises, although the rate of increase gradually diminishes. This phenomenon becomes more evident as the sequences lengthen (Figures~\ref{fig:serving} and~\ref{fig:serving3}), because the Dynamic SplitFuse strategy enables more fine-grained segmentation of longer sequences.
Regarding token latency, as the rate of requests escalates, both vLLM and MII show a steady increase in latency. 
Besides, the latency of the Dynamic SplitFuse strategy is observed to be higher when the GPU memory is limited (\ie 3090 and 4090).

%% file: sec-analysis.tex
\section{Fine-grained Modular Evaluation and Analysis} 
\label{sec:fine}

In this section, we conduct a both \emph{theoretical analysis} and \emph{practical evaluation} to quantify the time, floating point operations, memory read/write volumes, and arithmetic intensity of each module in LLaMA. This granular investigation provides a thorough understanding of the model's computational characteristics. Comparing these modules of Transformers and vLLM, we can derive insights into the optimization paths of current inference libraries and yield crucial guidance for future improvement.

\subsection{Theoretical Analysis}



In this section, we methodically dissect each operation within the LLaMA decoder layer, deriving theoretical formulas for the number of floating-point operations and the volume of memory reads/writes, as well as the resulting arithmetic intensity. This analysis is strictly limited to a single decoder layer; to extrapolate to real-world applications, one must multiply these findings by the total number of decoder layers. The outcomes of our analysis are detailed in Tables~\ref{tab-prefill-analysis} and~\ref{tab-decoding-analysis}.
In the following analysis, $b$ represents the batch size, $s$ represents the input sequence length, $h$ represents the hidden size, $h'$ represents the intermediate size of FFN module, $n$ represents the number of attention ``heads", and $d$ represents the size of each ``head" ($n$ and $d$ satisfying $h = n d$).

$\bullet$ \textbf{FLOP Analysis.}
First, let's analyze the \emph{prefill} phase:
For the MHA module, the three linear projections can be expressed as matrix multiplications (Equation~\ref{eq:qkv}), requiring $6 b s h^2$ FLOPs. The calculation of relative positional encoding (RoPE) involves 4 multiplications and 2 additions, requiring $6 b s h$ FLOPs.
Regarding the attention calculation (Equation~\ref{eq:attn}), the multiplication of matrix $\bm{Q}$ and matrix $\bm{K}$ requires $2 b s^2 h$ FLOPs. Dividing by $\sqrt{d}$ and calculating the softmax requires $4 b s^2 n$ FLOPs. Finally, multiplying with matrix $\bm{V}$ requires $2 b s^2 h$ FLOPs. Therefore, the attention calculation requires a total of $4 b s^2 h + 4 b s^2 n$ FLOPs.
The final linear transformation (Equation~\ref{eq:o}) in the MHA module also requires $2 b s h^2$ FLOPs.
For the FFN module (Equation~\ref{eq:ffn}), the initial two linear projections require $4 b s h h'$ FLOPs. The calculation of the activation function involves both multiplication and the Swish function, requiring $2 b s h'$ FLOPs. The final linear projection requires $2 b s h h'$ FLOPs.
The calculation of the RMS normalization (RMSNorm) and the residual networks requires $5 b s h'$ FLOPs.
For the \emph{decoding} phase, apart from the attention calculation, the FLOPs required for other parts can be obtained by substituting $s = 1$ into the corresponding formulas from the prefill phase. The FLOPs required for the attention calculation become $4 b s h + 4 b s n$.

\begin{table}[!t]
\resizebox{1 \textwidth}{!}{
\begin{tabular}{llllrrrrrr}
\toprule
                                                 & \textbf{}               & \textbf{}                       & \textbf{}                                                                   & \multicolumn{3}{c}{\textbf{Transformers}}    & \multicolumn{3}{c}{\textbf{vLLM}}            \\
                                                 & \textbf{FLOPs Form.}    & \textbf{I/O Form.}              & \textbf{A.I. Form.}                                                         & \textbf{Time} & \textbf{I/O} & \textbf{A.I.} & \textbf{Time} & \textbf{I/O} & \textbf{A.I.} \\ \midrule
$\bm{Q}, \bm{K}, \bm{V} = \bm{X} \bm{W}_{QKV}$   & $6 b s h^2$             & $\Theta(b s h + h^2)$           & $\Theta \left( \frac{1}{\frac{1}{h} + \frac{1}{bs}} \right)$                & 77.22         & 20.55        & 642.15        & 72.59         & 29.31        & 450.12        \\
$\bm{Q}, \bm{K} = \mathrm{RoPE}(\bm{Q}, \bm{K})$ & $6 b s h$               & $\Theta(b s h)$                 & $\Theta(1)$                                                                 & 32.79         & 17.66        & 0.18          & 5.34          & 3.47         & 0.93          \\
$\bm{O} = \mathrm{Attn}(\bm{Q}, \bm{K}, \bm{V})$ & $4 b s^2 h + 4 b s^2 n$ & $\Theta(b s^2 n + b s h)$       & $\Theta \left( \frac{1 + \frac{1}{d}}{\frac{1}{d} + \frac{1}{s}} \right)$   & 112.65        & 77.52        & 14.32         & 23.97         & 8.14         & 136.44        \\
$\bm{X} = \bm{O} \bm{W}_O$                       & $2 b s h^2$             & $\Theta(b s h + h^2)$           & $\Theta \left( \frac{1}{\frac{1}{h} + \frac{1}{bs}} \right)$                & 25.75         & 6.85         & 642.12        & 23.51         & 9.26         & 475.09        \\
$\bm{X} = \mathrm{Add}\&\mathrm{Norm}(\bm{X})$   & $5 b s h$               & $\Theta(b s h + h)$             & $\Theta \left( \frac{1}{1 + \frac{1}{bs}} \right)$                          & 18.47         & 19.80        & 0.14          & 4.99          & 3.40         & 0.79          \\
$\bm{G}, \bm{U} = \bm{X} [\bm{W}_G, \bm{W}_U]$   & $4 b s h h'$            & $\Theta(b s h + b s h' + h h')$ & $\Theta \left( \frac{1}{\frac{1}{h} + \frac{1}{h'} + \frac{1}{bs}} \right)$ & 119.91        & 37.52        & 630.02        & 128.37        & 53.02        & 445.85        \\
$\bm{D} = \mathrm{Swish}(\bm{G}) \odot \bm{U}$   & $2 b s h'$              & $\Theta(b s h')$                & $\Theta(1)$                                                                 & 9.23          & 13.60        & 0.21          & 9.15          & 8.11         & 0.36          \\
$\bm{X} = \bm{D} \bm{W}_D$                       & $2 b s h h'$            & $\Theta(b s h + b s h' + h h')$ & $\Theta \left( \frac{1}{\frac{1}{h} + \frac{1}{h'} + \frac{1}{bs}} \right)$ & 55.85         & 17.15        & 689.38        & 62.40         & 21.56        & 548.33        \\
$\bm{X} = \mathrm{Add}\&\mathrm{Norm}(\bm{X})$   & $5 b s h$               & $\Theta(b s h + h)$             & $\Theta \left( \frac{1}{1 + \frac{1}{bs}} \right)$                          & 18.47         & 19.80        & 0.14          & 4.99          & 3.40         & 0.79          \\ \bottomrule
\end{tabular}
}
\caption{
Theoretical and practical results of in prefill stage ($b=8, s=512$).
}
\label{tab-prefill-analysis}
\vspace{10pt}
\resizebox{1\textwidth}{!}{
\begin{tabular}{llllrrrrrr}
\toprule
                                                  & \textbf{}            & \textbf{}                        & \textbf{}                                                                     & \multicolumn{3}{c}{\textbf{Transformers}}    & \multicolumn{3}{c}{\textbf{vLLM}}            \\
                                                  & \textbf{FLOPs Form.} & \textbf{I/O Form.}               & \textbf{A.I. Form.}                                                           & \textbf{Time} & \textbf{I/O} & \textbf{A.I.} & \textbf{Time} & \textbf{I/O} & \textbf{A.I.} \\ \midrule
$\bm{q}, \bm{k}, \bm{v} = \bm{x} \bm{W}_{QKV}$    & $6 b h^2$            & $\Theta(b h + h^2)$              & $\Theta \left( \frac{1}{\frac{1}{h} + \frac{1}{b}} \right)$                   & 2.72          & 3.23         & 7.98          & 2.11          & 3.22         & 7.99          \\
$\bm{q}, \bm{k} = \mathrm{RoPE}(\bm{q}, \bm{k})$  & $6 b h$              & $\Theta(b h)$                    & $\Theta(1)$                                                                   & 2.66          & 0.03         & 0.24          & 0.31          & 0.00         & 1.48          \\
$\bm{K}, \bm{V} = \mathrm{Cache}(\bm{k}, \bm{v})$ & -                    & $\Theta(b h)$ or $\Theta(b s h)$ & -                                                                             & 10.89         & 3.46         & -             & 1.82          & 2.22         & -             \\
$\bm{o} = \mathrm{Attn}(\bm{q}, \bm{K}, \bm{V})$  & $4 b s h + 4 b s n$  & $\Theta(b s n + b s h + b h)$    & $\Theta \left( \frac{1 + \frac{1}{d}}{1 + \frac{1}{d} + \frac{1}{s}} \right)$ & 3.52          & 2.23         & 0.97          & 1.60          & 2.22         & 0.98          \\
$\bm{x} = \bm{o} \bm{W}_O$                        & $2 b h^2$            & $\Theta(b h + h^2)$              & $\Theta \left( \frac{1}{\frac{1}{h} + \frac{1}{b}} \right)$                   & 0.91          & 1.08         & 7.98          & 0.90          & 1.08         & 7.98          \\
$\bm{x} = \mathrm{Add}\&\mathrm{Norm}(\bm{x})$    & $5 b h$              & $\Theta(b h + h)$                & $\Theta \left( \frac{1}{1 + \frac{1}{b}} \right)$                             & 1.83          & 0.03         & 0.18          & 0.26          & 0.00         & 1.19          \\
$\bm{g}, \bm{u} = \bm{x} [\bm{W}_G, \bm{W}_U]$    & $4 b h h'$           & $\Theta(b h + b h' + h h')$      & $\Theta \left( \frac{1}{\frac{1}{h} + \frac{1}{h'} + \frac{1}{b}} \right)$    & 3.87          & 5.78         & 7.99          & 3.66          & 5.77         & 8.00          \\
$\bm{d} = \mathrm{Swish}(\bm{g}) \odot \bm{u}$    & $2 b h'$             & $\Theta(b h')$                   & $\Theta(1)$                                                                   & 0.27          & 0.02         & 0.33          & 0.42          & 0.01         & 0.50          \\
$\bm{x} = \bm{d} \bm{W}_D$                        & $2 b h h'$           & $\Theta(b h + b h' + h h')$      & $\Theta \left( \frac{1}{\frac{1}{h} + \frac{1}{h'} + \frac{1}{b}} \right)$    & 2.05          & 2.89         & 7.98          & 2.03          & 2.89         & 7.98          \\
$\bm{x} = \mathrm{Add}\&\mathrm{Norm}(\bm{x})$    & $5 b h$              & $\Theta(b h + h)$                & $\Theta \left( \frac{1}{1 + \frac{1}{b}} \right)$                             & 1.83          & 0.03         & 0.18          & 0.26          & 0.00         & 1.19          \\ \bottomrule
\end{tabular}
}
\caption{
Theoretical and practical results in decoding stage ($b=8, s=512$).
}
\label{tab-decoding-analysis}
\end{table}

$\bullet$ \textbf{MOPs Analysis.}
Due to the fact that matrix multiplication is calculated in blocks in practical operations, memory read and write volumes can only be expressed in the form of progressive complexity $\Theta$.
First, let's analyze the \emph{prefill} phase:
For the MHA module, the three linear projections can be expressed as matrix multiplications (Equation~\ref{eq:qkv}), requiring $\Theta( b s h )$ MOPs. The calculation of RoPE involves $\Theta( b s h )$ MOPs.
Regarding the attention calculation (Equation~\ref{eq:attn}), the multiplication of matrix $\bm{Q}$ and matrix $\bm{K}$ requires $\Theta( b s h + b s^2 n )$ MOPs. Dividing by $\sqrt{d}$ and calculating the softmax requires $\Theta( b s^2 n )$ MOPs. Finally, multiplying with matrix $\bm{V}$ requires $\Theta( b s h + b s^2 n )$ MOPs. Therefore, the attention calculation requires a total of $\Theta( b s h + b s^2 n )$ MOPs.
The final linear transformation (Equation~\ref{eq:o}) in the MHA module also requires $\Theta( b s h + b s h' + h h' )$ MOPs.
For the FFN module (Equation~\ref{eq:ffn}), the initial two linear projections $\Theta( b s h + b s h' + h h' )$ MOPs. The calculation of the activation function involves both multiplication and the Swish function, requiring $\Theta( b s h' )$ MOPs. The final linear projection requires $\Theta( b s h + b s h' + h h' )$ MOPs.
The calculation of the RMSNorm and the residual networks requires $\Theta( b s h + h )$ MOPs.
For the \emph{decoding} phase, apart from the attention calculation, the MOPs required for other parts can be obtained by substituting $s = 1$ into the corresponding formulas from the prefill phase. The MOPs required for the attention calculation become $\Theta(b s n + b s h + b h)$.

$\bullet$ \textbf{Arithmetic Intensity Analysis.}
Based on the analysis of FLOPs and MOPs, the arithmetic intensity of each module can be determined by dividing these two quantities. 
During the prefill stage, from the formulas in Table~\ref{tab-prefill-analysis}, it is evident that attention module exhibits the lowest arithmetic intensity, excluding components such as RoPE, RMSNorm, and residual networks.
During the decoding stage, the arithmetic intensity of each linear transformation is approximately $\Theta(b)$. However, the arithmetic intensity of the attention module is approximately $\Theta(1)$. 
Hence, optimizing the implementation of attention, RoPE, RMSNorm, and residual networks is crucial for reducing MOPs during the inference stage of LLMs, which leads to the development of FlashAttention~\citep{flash-attention} and PagedAttention~\citep{vllm}. Additionally, maximizing the batch size in the decoding stage is necessary to enhance the arithmetic intensity of linear transformations, which necessitates the advance of batching strategies~\citep{vllm,deepspeed-mii}.

\subsection{Evaluation Setup}





To accurately measure the execution time and memory read/write volume (MOPs) of various modules during real-world execution, we employ two tools: NVIDIA Nsight Compute CLI (NCU) and \texttt{torch.profile}. NCU is adept at quantifying the execution time and MOPs for individual CUDA kernels, while torch.profile offers detailed call stacks of CUDA kernels, enabling precise identification of specific modules.



\begin{figure*}[!t]
    \centering
    \includegraphics[width=0.98\textwidth]{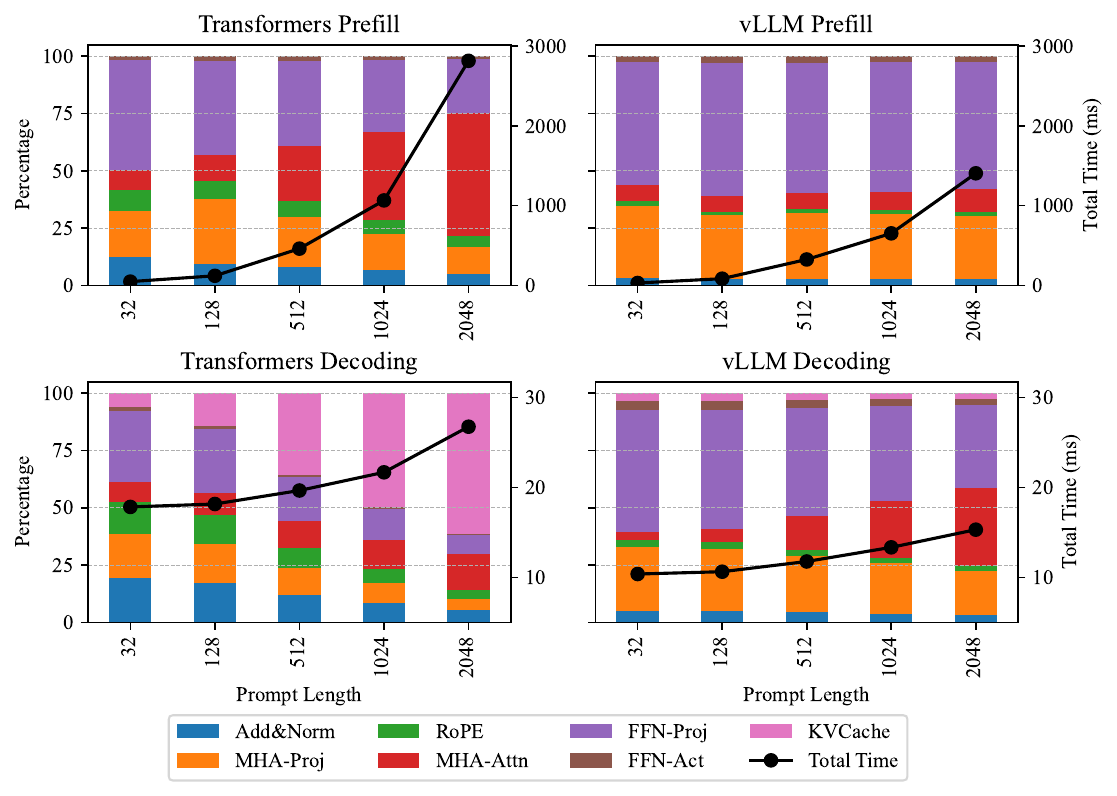}
    \caption{The distributions of total time and module time for both Transformers and vLLM libraries across different input lengths ranging from 32 to 2048 tokens.}
    \label{fig:fine-grained}
\end{figure*}

In the following experiments, we utilize simulated data with input lengths ranging from 32 to 2048 using a fixed batch size of $b = 8$ for Figure~\ref{fig:fine-grained}, Tables~\ref{exp-prefill-time-seq} and~\ref{exp-decoding-time-seq}. 
We also conduct experiments varying different batch sizes with $s = 1024$ in Figure~\ref{fig:fine-grained-batch}, Tables~\ref{exp-prefill-time-batch} and~\ref{exp-decoding-time-batch}, and experiments varying different hardware with $b = 8, s = 512$ in Tables~\ref{exp-prefill-time-gpu} and ~\ref{exp-decoding-time-gpu}.
For each experiment, we employ both Transformers and vLLM libraries to generate two tokens using LLaMA-2 (7B) each on A800 GPU. This allows for execution of the prefill stage and the decoding stage once within each library. 


\subsection{Evaluation Results}

Firstly, our practical time consuming results are consistent with our theoretical analysis results in Tables~\ref{tab-prefill-analysis} and~\ref{tab-decoding-analysis}. 
Thus, we can estimate the runtime for each module during the prefill and decoding phases. For compute-bound operations (\eg the linear transformation of MHA in the prefill phase), the estimation primarily relies on the number of FLOPs, represented as $bsh^2$. For memory-bound operations, runtime is primarily influenced by the volume of I/O operations. The corresponding estimation equations are detailed below:
\begin{align} \label{eq:time-formula}
T_{\text{prefill}} & = \alpha \underbrace{b s h^2 l}_{\text{MHA Proj.}} + \beta \underbrace{b s h h' l}_{\text{FFN Proj.}} + \gamma \underbrace{b s^2 n l}_{\text{Attn.}} + \eta \underbrace{b s h l}_{\text{RoPE, Norm, Res., Attn.}} + \lambda \underbrace{b s h' l}_{\text{FFN Act.}} + \mu, \\
T_{\text{decoding}} & = \phi \underbrace{b s h l}_{\text{KV Cache, Attn.}} + \psi \underbrace{b s n l}_{\text{KV Cache}} + \omega \underbrace{b h l}_{\text{KV Cache, Attn.}} + \nu,
\end{align}
where $\alpha, \beta, \gamma, \eta, \lambda, \mu, \phi, \psi, \omega, \nu$ are the coefficients of different items. We can determine them through linear regression based on our experimental data, as presented in Table \ref{exp-time-formula}.




Secondly, the attention module is the bottleneck during the prefill and decoding stage from the results in Figure~\ref{fig:fine-grained}. 
Notably, during the prefill phase, the conventional attention mechanism emerges as the primary bottleneck, particularly as the input length escalates. To address this challenge, the integration of FlashAttention~\citep{flash-attention} presents an effective optimization strategy.
Conversely, in the decoding phase, inadequate management of the KV cache can result in the update of the KV cache emerging as the principal bottleneck with increasing input lengths. vLLM employs block management techniques for KV cache and PagedAttention~\citep{vllm} mechanisms to streamline KV cache updates and attention calculations, contributing to enhanced efficiency in decoding tasks.


Third, batching strategies are shown to be effective for increasing arithmetic intensity during the decoding stage. According to the formulas presented in Table~\ref{tab-decoding-analysis}, it is evident that all operations are memory-bound during decoding due to the low arithmetic intensity. For operations such as linear transformations and activations, increasing the batch size can enhance arithmetic intensity. Notably, even with larger batch sizes and input lengths, the processing time remains nearly consistent for these operations, as indicated in Tables~\ref{exp-decoding-time-batch} and~\ref{exp-decoding-time-seq}. This consistency suggests that we can execute more FLOPs within a similar timeframe. Such findings support the use of strategies such as continuous batching~\cite{vllm} and Dynamic SplitFuse~\cite{deepspeed-mii} to boost arithmetic intensity and thereby increase the overall token throughput.


In addition, CUDA kernel fusion also plays a significant role in improving decoding efficiency. The vLLM library features specially designed CUDA kernels tailored for operations such as RoPE, Swish, and RMSNorm. In contrast to the Transformers library, which exhibits the same arithmetic intensity complexity (as shown in Tables~\ref{tab-prefill-analysis} and~\ref{tab-decoding-analysis}), vLLM refines the implementation of these operations to optimize memory access patterns and reduce execution time, as illustrated in Figure~\ref{fig:fine-grained}.



Finally, it is evident that linear transformation operations (\ie MHA projection and FFN projection) still occupy a substantial portion of time during both the prefill and decoding phases.
As shown in Figure~\ref{fig:fine-grained}, after various vLLM optimizations, linear transformations comprise the most time-consuming elements when the sequence length is short and they also account for over 50\% of the total time as the sequence length increases. Although optimizing matrix multiplication presents inherent challenges, it offers a promising path for future inference enhancements.



%% file: sec-rel.tex
\section{Related Work}









$\bullet$ \textbf{System Optimization.}
There are many optimization algorithms for inference in large language models. To address the low efficiency issue of multi-head attention calculation, \emph{FlashAttention}~\citep{flash-attention,flash-attention-2} leverages optimization strategies employed in GPU-based matrix multiplication. This involves partitioning the matrices $\bm{Q}$, $\bm{K}$, and $\bm{V}$ and directly computing the resultant matrix $\bm{O}$. By reducing the frequency of memory accesses, this optimization technique increases the arithmetic intensity and improves the efficiency of the attention module.
To optimize the management of the KV cache memory in decoding phase, vLLM proposes \emph{PagedAttention}~\citep{vllm}. This mechanism effectively mitigates the frequent update requirement of the KV cache and reduces memory fragmentation, leading to improved overall efficiency.
In practical applications, it is often necessary to handle multiple requests concurrently. While the current model architecture supports batch inference, a straightforward implementation requires completing all requests within a batch before initiating the next batch.
To address this limitation, researchers have proposed batching strategy such as \emph{continuous batching}~\citep{vllm}, \emph{inflight batching}~\citep{trt-llm} and \emph{Dynamic SplitFuse}~\citep{deepspeed-mii}.
Their strategy involves immediately substituting completed requests with new ones, eliminating the need for padding tokens. This streamlined processing pipeline enhances throughput and efficiency by ensuring continuous computation without idle periods.

$\bullet$ \textbf{Inference Libraries.}
The Transformers~\citep{wolftransformers} library is a commonly used library in the field of natural language processing, providing code and archive points for many common models, making it convenient for users to use.
TGI~\citep{TGI} is a library developed by HuggingFace for further optimization of inference based on the Transformers.
vLLM~\citep{vllm} mainly adopts targeted optimization strategies in terms of decoding efficiency, significantly improving the utilization efficiency of KV Cache by paging storage and combining with PagedAttention technology.
DeepSpeed-MII has introduced Dynamic-SplitFuse technology to fully tap into the computing potential of GPUs. This technology achieves an increase in batch data and decoding throughput by splitting input prompts into multiple sub blocks and fusing requests for full decoding and incremental decoding.
TensorRT-LLM is developed by Nvidia, which has been further optimized based on the previous FasterTransformer~\citep{FasterTransformer} library, improving the efficiency of running large models on Nvidia GPUs.
Llama.cpp is entirely based on C/C++ implementation, with good cross platform compatibility and the ability to run on various computing devices. 
Other code libraries such as LightLLM~\citep{lightllm}, LMDeploy~\citep{2023lmdeploy}, StreamLLM~\citep{xiao2023streamingllm}, and Inferflow~\citep{shi2024inferflow}  have made different optimization implementations for inference in LLMs.















%% file: sec-con.tex
\section{Conclusion}

In this work, we introduced a comprehensive benchmark that can encompass diverse task scenarios for the evaluation of various libraries. 
We integrated various common experimental settings in our framework, to provide a useful testbed for evaluating inference efficiency related libraries.
Based on it, we proposed a detailed formula for analyzing the complexity of each component from LLaMA, which involves metrics such as FLOPs, MOPs, and arithmetic intensity.
It can delineate the decoding bottlenecks in the inadequacy of memory bandwidth, and has been validated in our experiments.
Besides, widely-used strategies and toolkits such as FlashAttention, PagedAttention and CUDA kernel fusion demonstrated the mitigation of memory access constraints is helpful to enhance the inference efficiency. 
It is anticipated that our findings will offer valuable insights for the advancement of future inference algorithms and libraries.




%% file: sec-app.tex
\clearpage
\appendix

\section{Appendix}


\begin{table}[H]
\centering
\begin{tabular}{llll}
\toprule
\textbf{}                 & \textbf{3090}     & \textbf{4090}     & \textbf{A800}     \\ \midrule
\textbf{Memory Size (GB)} & 24                & 24                & 80                \\
\textbf{Bandwidth (GB/s)} & 936               & 1008              & 2039              \\
\textbf{BF16 TFLOPs}      & 71                & 165.2             & 312               \\ \bottomrule
\end{tabular}
\caption{The details of three hardwares.}
\label{tab-hardwares}
\end{table}

\begin{table}[H]
\centering
\begin{tabular}{lccrrrrr}
\toprule
\textbf{Data}                   & \textbf{Hardware} & \textbf{Model Size} & \textbf{TRF} & \textbf{vLLM} & \textbf{MII} & \textbf{TRT} & \textbf{L.CPP} \\ \midrule
\multirow{4}{*}{\textbf{S2S}}   & \textbf{3090}     & \textbf{7B}         & 272.26       & 1121.62       & 1072.79      & 356.72       & 406.95         \\
                                & \textbf{4090}     & \textbf{7B}         & 379.65       & 1924.46       & 1094.31      & 445.09       & 238.98         \\
                                & \textbf{A800}     & \textbf{7B}         & 413.12       & 2167.77       & 1596.45      & 628.72       & 479.58         \\
                                & \textbf{A800}     & \textbf{13B}        & 147.75       & 1515.21       & 1235.56      & 342.50       & 496.52         \\ \midrule
\multirow{4}{*}{\textbf{S2L}}   & \textbf{3090}     & \textbf{7B}         & 102.74       & 860.99        & 610.83       & 358.96       & 223.05         \\
                                & \textbf{4090}     & \textbf{7B}         & 87.65        & 1154.60       & 674.17       & 399.63       & 210.57         \\
                                & \textbf{A800}     & \textbf{7B}         & 101.74       & 2771.10       & 808.82       & 634.09       & 280.53         \\
                                & \textbf{A800}     & \textbf{13B}        & 77.74        & 1797.52       & 562.07       & 339.68       & 326.98         \\ \midrule
\multirow{4}{*}{\textbf{L2S}}   & \textbf{3090}     & \textbf{7B}         & 48.18        & 124.29        & 112.85       & 98.21        & 58.71          \\
                                & \textbf{4090}     & \textbf{7B}         & 64.78        & 205.32        & 166.64       & 180.57       & 46.67          \\
                                & \textbf{A800}     & \textbf{7B}         & 73.92        & 331.19        & 232.79       & 269.58       & 15.92          \\
                                & \textbf{A800}     & \textbf{13B}        & 30.68        & 254.89        & 96.63        & 147.15       & 13.51          \\ \midrule
\multirow{2}{*}{\textbf{S-16k}} & \textbf{A800}     & \textbf{7B}         & 25.31        & 214.03        & 185.15       & 122.32       & 35.40          \\
                                & \textbf{A800}     & \textbf{13B}        & 17.01        & 115.58        & 90.22        & 38.03        & 27.80          \\ \bottomrule
\end{tabular}
\caption{
The token throughput (token/s) for batch inference using LLaMA-2 (7B) and (13B).
}
\label{exp-batch-throughput}
\end{table}

\begin{figure}[H]
    \centering
    \includegraphics[width=1\textwidth]{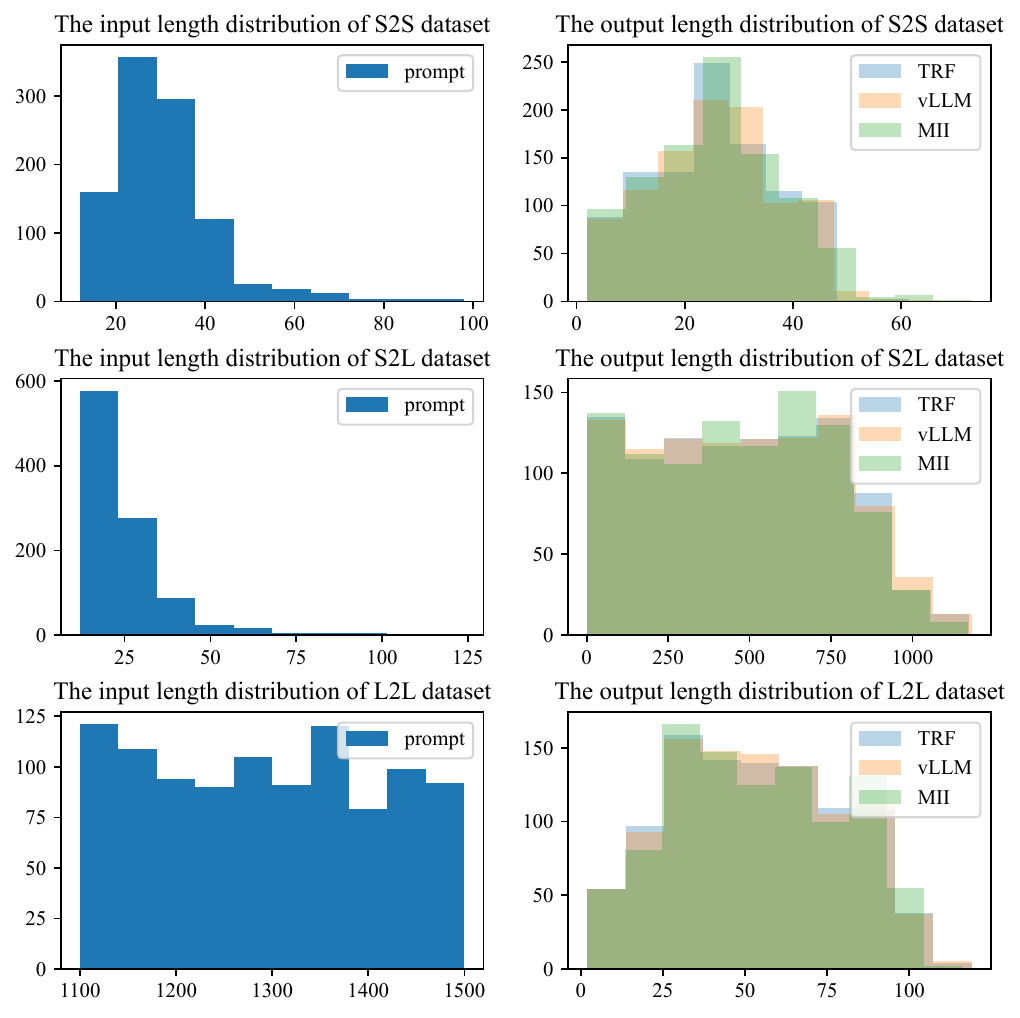}
    \caption{The distribution of input length and output length of three datasets.}
    \label{fig:length}
\end{figure}

\begin{figure}[H]
    \centering
    \includegraphics[width=1\textwidth]{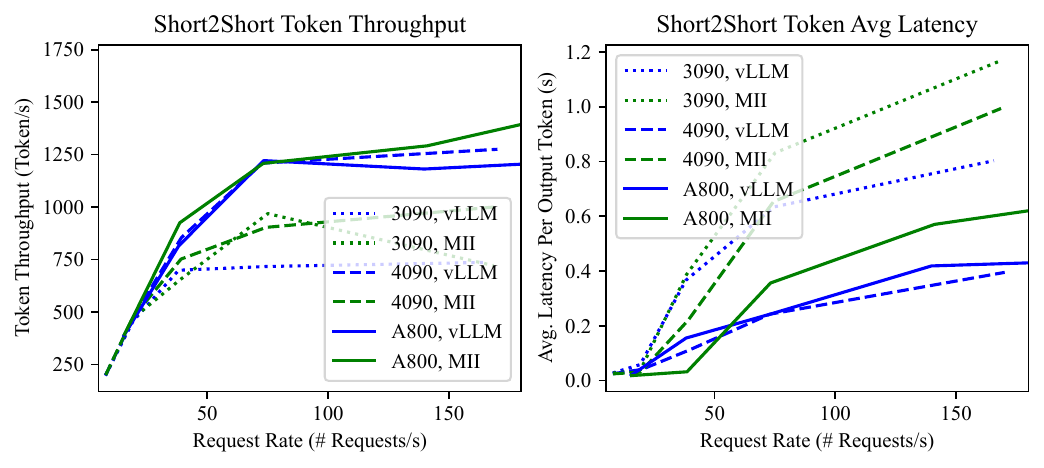}
    \caption{
        The throughput and latency for serving inference of vLLM and MII using LLaMA-2 (7B) under different request frequencies on the Short2Short dataset.
    }
    \label{fig:serving2}
\end{figure}

\begin{figure}[H]
    \centering
    \includegraphics[width=1\textwidth]{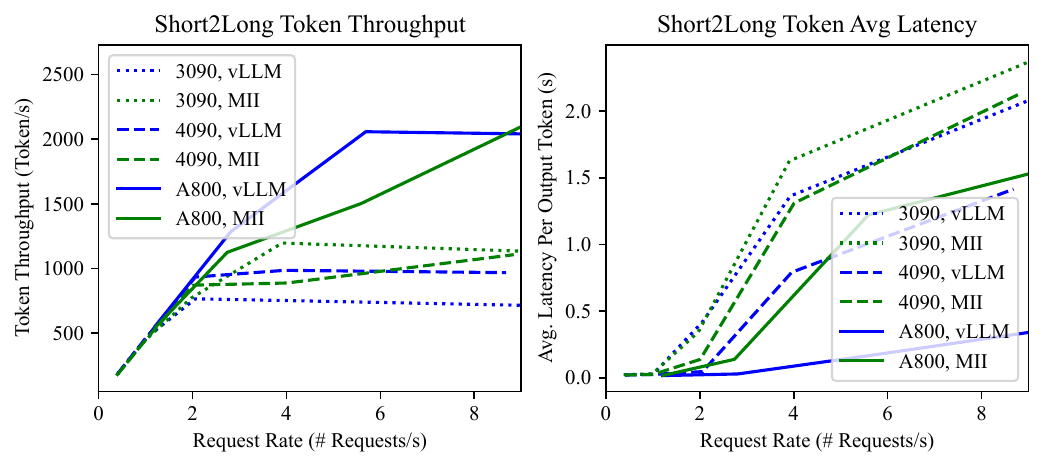}
    \caption{
        The throughput and latency for serving inference of vLLM and MII using LLaMA-2 (7B) under different request frequencies on the Short2Long dataset.
    }
    \label{fig:serving3}
\end{figure}

\begin{figure}[H]
    \centering
    \includegraphics[width=1\textwidth]{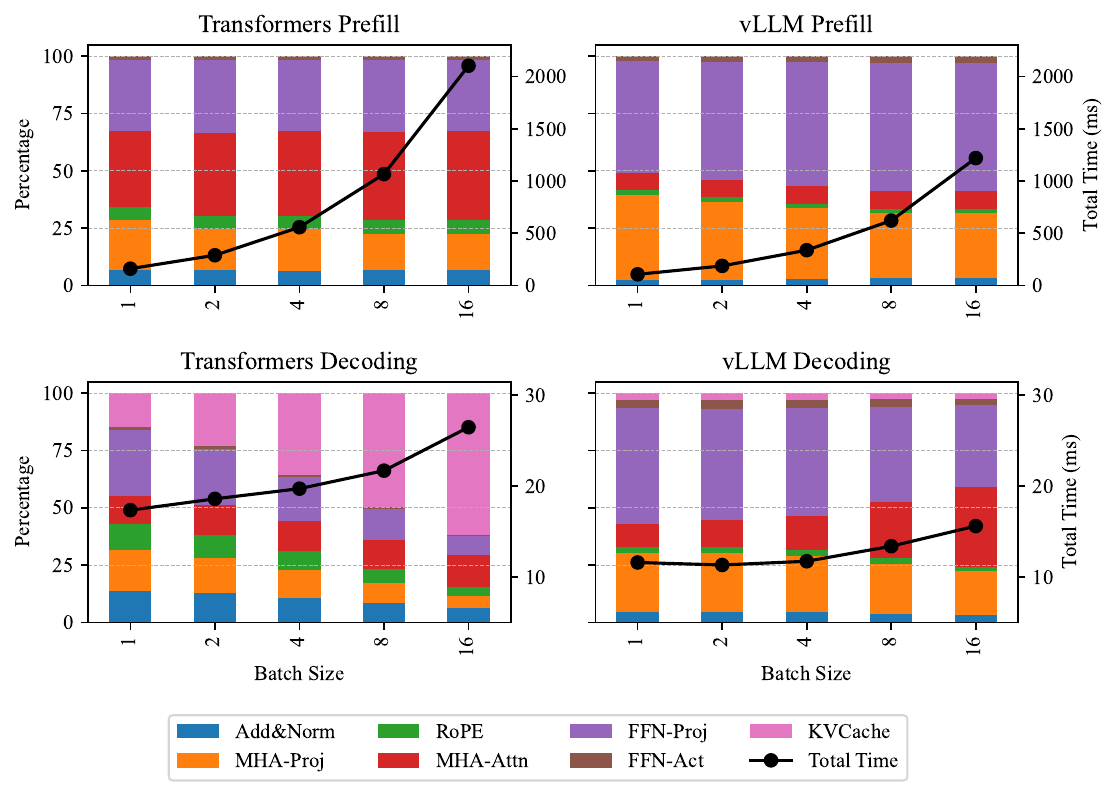}
    \caption{The distribution of total time and module time for both Transformers and vLLM libraries across different batch size ranging from 1 to 16.}
    \label{fig:fine-grained-batch}
\end{figure}

\begin{table}[H]
\centering
\begin{tabular}{llrrrrr}
\toprule
                               &                                                           & \multicolumn{1}{c}{\textbf{$s=32$}} & \multicolumn{1}{c}{\textbf{$s=128$}} & \multicolumn{1}{c}{\textbf{$s=512$}} & \multicolumn{1}{c}{\textbf{$s=1024$}} & \multicolumn{1}{c}{\textbf{$s=2048$}} \\ \midrule
\multirow{9}{*}{\textbf{TRF}}  & \textbf{$\bm{Q}, \bm{K}, \bm{V} = \bm{X} \bm{W}_{QKV}$}   & 7.33                                & 26.27                                & 77.22                                & 129.09                                & 256.14                                \\
                               & \textbf{$\bm{Q}, \bm{K} = \mathrm{RoPE}(\bm{Q}, \bm{K})$} & 4.63                                & 9.94                                 & 32.79                                & 63.79                                 & 125.63                                \\
                               & \textbf{$\bm{O} = \mathrm{Attn}(\bm{Q}, \bm{K}, \bm{V})$} & 4.15                                & 14.05                                & 112.65                               & 415.68                                & 1544.94                               \\
                               & \textbf{$\bm{X} = \bm{O} \bm{W}_O$}                       & 2.45                                & 8.77                                 & 25.75                                & 43.05                                 & 85.39                                 \\
                               & \textbf{$\bm{X} = \mathrm{Add}\&\mathrm{Norm}(\bm{X})$}   & 3.08                                & 5.68                                 & 18.47                                & 36.32                                 & 71.36                                 \\
                               & \textbf{$\bm{G}, \bm{U} = \bm{X} [\bm{W}_G, \bm{W}_U]$}   & 23.73                               & 50.92                                & 175.76                               & 341.57                                & 665.18                                \\
                               & \textbf{$\bm{D} = \mathrm{Swish}(\bm{G}) \odot \bm{U}$}    & 0.79                                & 2.38                                 & 9.23                                 & 18.42                                 & 36.80                                 \\
                               & \textbf{$\bm{X} = \bm{D} \bm{W}_D$}                       & 6.19                                & 16.18                                & 55.85                                & 111.27                                & 221.92                                \\
                               & \textbf{$\bm{X} = \mathrm{Add}\&\mathrm{Norm}(\bm{X})$}   & 3.08                                & 5.68                                 & 18.47                                & 36.32                                 & 71.36                                 \\ \midrule
\multirow{9}{*}{\textbf{vLLM}} & \textbf{$\bm{Q}, \bm{K}, \bm{V} = \bm{X} \bm{W}_{QKV}$}   & 7.26                                & 17.42                                & 72.59                                & 144.55                                & 299.84                                \\
                               & \textbf{$\bm{Q}, \bm{K} = \mathrm{RoPE}(\bm{Q}, \bm{K})$} & 0.67                                & 1.38                                 & 5.34                                 & 10.33                                 & 20.97                                 \\
                               & \textbf{$\bm{O} = \mathrm{Attn}(\bm{Q}, \bm{K}, \bm{V})$} & 2.21                                & 6.00                                 & 23.97                                & 54.08                                 & 147.36                                \\
                               & \textbf{$\bm{X} = \bm{O} \bm{W}_O$}                       & 2.73                                & 6.98                                 & 23.51                                & 46.22                                 & 99.65                                 \\
                               & \textbf{$\bm{X} = \mathrm{Add}\&\mathrm{Norm}(\bm{X})$}   & 0.52                                & 1.25                                 & 4.99                                 & 9.72                                  & 19.42                                 \\
                               & \textbf{$\bm{G}, \bm{U} = \bm{X} [\bm{W}_G, \bm{W}_U]$}   & 11.07                               & 32.18                                & 128.37                               & 255.03                                & 527.56                                \\
                               & \textbf{$\bm{D} = \mathrm{Swish}(\bm{G}) \odot \bm{U}$}    & 0.83                                & 2.39                                 & 9.15                                 & 17.97                                 & 36.36                                 \\
                               & \textbf{$\bm{X} = \bm{D} \bm{W}_D$}                       & 5.96                                & 18.66                                & 62.40                                & 123.06                                & 271.20                                \\
                               & \textbf{$\bm{X} = \mathrm{Add}\&\mathrm{Norm}(\bm{X})$}   & 0.52                                & 1.25                                 & 4.99                                 & 9.72                                  & 19.42                                 \\ \bottomrule
\end{tabular}
\caption{
Detailed running time (ms) of Transformers and vLLM when varying sequence length in the prefill stage ($b = 8$).
}
\label{exp-prefill-time-seq}
\end{table}

\begin{table}[H]
\centering
\begin{tabular}{llrrrrr}
\toprule
                                &                                                            & \multicolumn{1}{c}{\textbf{$s=32$}} & \multicolumn{1}{c}{\textbf{$s=128$}} & \multicolumn{1}{c}{\textbf{$s=512$}} & \multicolumn{1}{c}{\textbf{$s=1024$}} & \multicolumn{1}{c}{\textbf{$s=2048$}} \\ \midrule
\multirow{10}{*}{\textbf{TRF}}  & \textbf{$\bm{q}, \bm{k}, \bm{v} = \bm{x} \bm{W}_{QKV}$}    & 2.73                                & 2.72                                 & 2.72                                 & 2.72                                  & 2.70                                  \\
                                & \textbf{$\bm{q}, \bm{k} = \mathrm{RoPE}(\bm{q}, \bm{k})$}  & 2.68                                & 2.68                                 & 2.66                                 & 2.67                                  & 2.67                                  \\
                                & \textbf{$\bm{K}, \bm{V} = \mathrm{Cache}(\bm{k}, \bm{v})$} & 1.18                                & 3.05                                 & 10.89                                & 21.51                                 & 42.70                                 \\
                                & \textbf{$\bm{o} = \mathrm{Attn}(\bm{q}, \bm{K}, \bm{V})$}  & 1.65                                & 2.01                                 & 3.52                                 & 5.48                                  & 10.70                                 \\
                                & \textbf{$\bm{x} = \bm{o} \bm{W}_O$}                        & 0.91                                & 0.91                                 & 0.91                                 & 0.92                                  & 0.90                                  \\
                                & \textbf{$\bm{x} = \mathrm{Add}\&\mathrm{Norm}(\bm{x})$}    & 1.83                                & 1.82                                 & 1.83                                 & 1.84                                  & 1.80                                  \\
                                & \textbf{$\bm{g}, \bm{u} = \bm{x} [\bm{W}_G, \bm{W}_U]$}    & 3.88                                & 3.87                                 & 3.87                                 & 3.88                                  & 3.87                                  \\
                                & \textbf{$\bm{d} = \mathrm{Swish}(\bm{g}) \odot \bm{u}$}          & 0.27                                & 0.27                                 & 0.27                                 & 0.27                                  & 0.27                                  \\
                                & \textbf{$\bm{x} = \bm{d} \bm{W}_D$}                        & 2.05                                & 2.05                                 & 2.05                                 & 2.05                                  & 2.04                                  \\
                                & \textbf{$\bm{x} = \mathrm{Add}\&\mathrm{Norm}(\bm{x})$}    & 1.83                                & 1.82                                 & 1.83                                 & 1.84                                  & 1.80                                  \\ \midrule
\multirow{10}{*}{\textbf{vLLM}} & \textbf{$\bm{q}, \bm{k}, \bm{v} = \bm{x} \bm{W}_{QKV}$}    & 2.11                                & 2.11                                 & 2.11                                 & 2.11                                  & 2.11                                  \\
                                & \textbf{$\bm{q}, \bm{k} = \mathrm{RoPE}(\bm{q}, \bm{k})$}  & 0.32                                & 0.31                                 & 0.31                                 & 0.31                                  & 0.31                                  \\
                                & \textbf{$\bm{K}, \bm{V} = \mathrm{Cache}(\bm{k}, \bm{v})$} & 0.38                                & 0.38                                 & 0.38                                 & 0.38                                  & 0.38                                  \\
                                & \textbf{$\bm{o} = \mathrm{Attn}(\bm{q}, \bm{K}, \bm{V})$}  & 0.40                                & 0.65                                 & 1.81                                 & 3.40                                  & 5.32                                  \\
                                & \textbf{$\bm{x} = \bm{o} \bm{W}_O$}                        & 0.90                                & 0.89                                 & 0.89                                 & 0.89                                  & 0.90                                  \\
                                & \textbf{$\bm{x} = \mathrm{Add}\&\mathrm{Norm}(\bm{x})$}    & 0.27                                & 0.27                                 & 0.26                                 & 0.26                                  & 0.27                                  \\
                                & \textbf{$\bm{g}, \bm{u} = \bm{x} [\bm{W}_G, \bm{W}_U]$}    & 3.67                                & 3.68                                 & 3.66                                 & 3.66                                  & 3.68                                  \\
                                & \textbf{$\bm{d} = \mathrm{Swish}(\bm{g}) \odot \bm{u}$}          & 0.42                                & 0.42                                 & 0.42                                 & 0.42                                  & 0.42                                  \\
                                & \textbf{$\bm{x} = \bm{d} \bm{W}_D$}                        & 2.03                                & 2.03                                 & 2.03                                 & 2.02                                  & 2.03                                  \\
                                & \textbf{$\bm{x} = \mathrm{Add}\&\mathrm{Norm}(\bm{x})$}    & 0.27                                & 0.27                                 & 0.26                                 & 0.26                                  & 0.27                                  \\ \bottomrule
\end{tabular}
\caption{
Detailed running time (ms) of Transformers and vLLM when varying sequence length in the decoding stage ($b = 8$).
}
\label{exp-decoding-time-seq}
\end{table}

\begin{table}[H]
\centering
\begin{tabular}{llrrrrr}
\toprule
                               &                                                           & \multicolumn{1}{c}{\textbf{$b=1$}} & \multicolumn{1}{c}{\textbf{$b=2$}} & \multicolumn{1}{c}{\textbf{$b=4$}} & \multicolumn{1}{c}{\textbf{$b=8$}} & \multicolumn{1}{c}{\textbf{$b=16$}} \\ \midrule
\multirow{9}{*}{\textbf{TRF}}  & \textbf{$\bm{Q}, \bm{K}, \bm{V} = \bm{X} \bm{W}_{QKV}$}   & 26.28                              & 39.20                              & 77.26                              & 129.09                             & 256.15                              \\
                               & \textbf{$\bm{Q}, \bm{K} = \mathrm{RoPE}(\bm{Q}, \bm{K})$} & 9.64                               & 17.48                              & 32.82                              & 63.79                              & 125.74                              \\
                               & \textbf{$\bm{O} = \mathrm{Attn}(\bm{Q}, \bm{K}, \bm{V})$} & 53.97                              & 105.54                             & 209.38                             & 415.68                             & 827.96                              \\
                               & \textbf{$\bm{X} = \bm{O} \bm{W}_O$}                       & 8.77                               & 13.08                              & 25.76                              & 43.05                              & 85.40                               \\
                               & \textbf{$\bm{X} = \mathrm{Add}\&\mathrm{Norm}(\bm{X})$}   & 5.66                               & 9.77                               & 18.46                              & 36.32                              & 71.41                               \\
                               & \textbf{$\bm{G}, \bm{U} = \bm{X} [\bm{W}_G, \bm{W}_U]$}   & 50.93                              & 94.01                              & 175.78                             & 341.57                             & 665.21                              \\
                               & \textbf{$\bm{D} = \mathrm{Swish}(\bm{G}) \odot \bm{U}$}    & 2.36                               & 4.64                               & 9.23                               & 18.42                              & 36.82                               \\
                               & \textbf{$\bm{X} = \bm{D} \bm{W}_D$}                       & 16.18                              & 33.54                              & 55.85                              & 111.27                             & 221.92                              \\
                               & \textbf{$\bm{X} = \mathrm{Add}\&\mathrm{Norm}(\bm{X})$}   & 5.66                               & 9.77                               & 18.46                              & 36.32                              & 71.41                               \\ \midrule
\multirow{9}{*}{\textbf{vLLM}} & \textbf{$\bm{Q}, \bm{K}, \bm{V} = \bm{X} \bm{W}_{QKV}$}   & 33.58                              & 53.01                              & 85.13                              & 138.82                             & 271.39                              \\
                               & \textbf{$\bm{Q}, \bm{K} = \mathrm{RoPE}(\bm{Q}, \bm{K})$} & 2.58                               & 3.77                               & 6.22                               & 10.14                              & 20.18                               \\
                               & \textbf{$\bm{O} = \mathrm{Attn}(\bm{Q}, \bm{K}, \bm{V})$} & 8.16                               & 14.05                              & 26.04                              & 49.95                              & 97.63                               \\
                               & \textbf{$\bm{X} = \bm{O} \bm{W}_O$}                       & 6.94                               & 12.37                              & 23.01                              & 44.28                              & 86.57                               \\
                               & \textbf{$\bm{X} = \mathrm{Add}\&\mathrm{Norm}(\bm{X})$}   & 1.21                               & 2.43                               & 4.86                               & 9.76                               & 19.53                               \\
                               & \textbf{$\bm{G}, \bm{U} = \bm{X} [\bm{W}_G, \bm{W}_U]$}   & 34.39                              & 65.60                              & 124.65                             & 243.48                             & 480.48                              \\
                               & \textbf{$\bm{D} = \mathrm{Swish}(\bm{G}) \odot \bm{U}$}    & 2.37                               & 4.54                               & 8.91                               & 17.62                              & 35.04                               \\
                               & \textbf{$\bm{X} = \bm{D} \bm{W}_D$}                       & 18.59                              & 33.70                              & 63.38                              & 115.57                             & 224.88                              \\
                               & \textbf{$\bm{X} = \mathrm{Add}\&\mathrm{Norm}(\bm{X})$}   & 1.21                               & 2.43                               & 4.86                               & 9.76                               & 19.53                               \\ \bottomrule
\end{tabular}
\caption{
Detailed running time (ms) of Transformers and vLLM when varying batch size in the prefill stage ($s = 1024$).
}
\label{exp-prefill-time-batch}
\end{table}

\begin{table}[H]
\centering

\begin{tabular}{llrrrrr}
\toprule
                                &                                                            & \multicolumn{1}{c}{\textbf{$b=1$}} & \multicolumn{1}{c}{\textbf{$b=2$}} & \multicolumn{1}{c}{\textbf{$b=4$}} & \multicolumn{1}{c}{\textbf{$b=8$}} & \multicolumn{1}{c}{\textbf{$b=16$}} \\ \midrule
\multirow{10}{*}{\textbf{TRF}}  & \textbf{$\bm{q}, \bm{k}, \bm{v} = \bm{x} \bm{W}_{QKV}$}    & 2.70                               & 2.71                               & 2.72                               & 2.72                               & 2.72                                \\
                                & \textbf{$\bm{q}, \bm{k} = \mathrm{RoPE}(\bm{q}, \bm{k})$}  & 2.34                               & 2.45                               & 2.61                               & 2.67                               & 2.72                                \\
                                & \textbf{$\bm{K}, \bm{V} = \mathrm{Cache}(\bm{k}, \bm{v})$} & 2.98                               & 5.59                               & 10.91                              & 21.51                              & 42.77                               \\
                                & \textbf{$\bm{o} = \mathrm{Attn}(\bm{q}, \bm{K}, \bm{V})$}  & 2.47                               & 3.22                               & 3.97                               & 5.48                               & 9.50                                \\
                                & \textbf{$\bm{x} = \bm{o} \bm{W}_O$}                        & 0.91                               & 0.91                               & 0.91                               & 0.92                               & 0.92                                \\
                                & \textbf{$\bm{x} = \mathrm{Add}\&\mathrm{Norm}(\bm{x})$}    & 1.40                               & 1.57                               & 1.65                               & 1.84                               & 2.18                                \\
                                & \textbf{$\bm{g}, \bm{u} = \bm{x} [\bm{W}_G, \bm{W}_U]$}    & 3.83                               & 3.85                               & 3.88                               & 3.88                               & 3.89                                \\
                                & \textbf{$\bm{d} = \mathrm{Swish}(\bm{g}) \odot \bm{u}$}          & 0.25                               & 0.25                               & 0.26                               & 0.27                               & 0.28                                \\
                                & \textbf{$\bm{x} = \bm{d} \bm{W}_D$}                        & 2.05                               & 2.05                               & 2.05                               & 2.05                               & 2.06                                \\
                                & \textbf{$\bm{x} = \mathrm{Add}\&\mathrm{Norm}(\bm{x})$}    & 1.40                               & 1.57                               & 1.65                               & 1.84                               & 2.18                                \\ \midrule
\multirow{10}{*}{\textbf{vLLM}} & \textbf{$\bm{q}, \bm{k}, \bm{v} = \bm{x} \bm{W}_{QKV}$}    & 2.18                               & 2.10                               & 2.10                               & 2.11                               & 2.14                                \\
                                & \textbf{$\bm{q}, \bm{k} = \mathrm{RoPE}(\bm{q}, \bm{k})$}  & 0.30                               & 0.31                               & 0.31                               & 0.31                               & 0.32                                \\
                                & \textbf{$\bm{K}, \bm{V} = \mathrm{Cache}(\bm{k}, \bm{v})$} & 0.35                               & 0.37                               & 0.37                               & 0.38                               & 0.39                                \\
                                & \textbf{$\bm{o} = \mathrm{Attn}(\bm{q}, \bm{K}, \bm{V})$}  & 1.22                               & 1.37                               & 1.78                               & 3.41                               & 5.56                                \\
                                & \textbf{$\bm{x} = \bm{o} \bm{W}_O$}                        & 0.89                               & 0.89                               & 0.89                               & 0.89                               & 0.90                                \\
                                & \textbf{$\bm{x} = \mathrm{Add}\&\mathrm{Norm}(\bm{x})$}    & 0.27                               & 0.27                               & 0.26                               & 0.26                               & 0.26                                \\
                                & \textbf{$\bm{g}, \bm{u} = \bm{x} [\bm{W}_G, \bm{W}_U]$}    & 4.02                               & 3.67                               & 3.66                               & 3.67                               & 3.67                                \\
                                & \textbf{$\bm{d} = \mathrm{Swish}(\bm{g}) \odot \bm{u}$}          & 0.40                               & 0.41                               & 0.42                               & 0.42                               & 0.43                                \\
                                & \textbf{$\bm{x} = \bm{d} \bm{W}_D$}                        & 2.03                               & 2.02                               & 2.02                               & 2.03                               & 2.03                                \\
                                & \textbf{$\bm{x} = \mathrm{Add}\&\mathrm{Norm}(\bm{x})$}    & 0.27                               & 0.27                               & 0.26                               & 0.26                               & 0.26                                \\ \bottomrule
\end{tabular}
\caption{
Detailed running time (ms) of Transformers and vLLM when varying batch size in the decoding stage ($s = 1024$).
}
\label{exp-decoding-time-batch}
\end{table}

\begin{table}[H]
\centering
\begin{tabular}{llrrr}
\toprule
                               &                                                           & \multicolumn{1}{c}{\textbf{3090}} & \multicolumn{1}{c}{\textbf{4090}} & \multicolumn{1}{c}{\textbf{A800}} \\ \midrule
\multirow{9}{*}{\textbf{TRF}}  & \textbf{$\bm{Q}, \bm{K}, \bm{V} = \bm{X} \bm{W}_{QKV}$}   & 243.31                            & 115.54                            & 77.22                             \\
                               & \textbf{$\bm{Q}, \bm{K} = \mathrm{RoPE}(\bm{Q}, \bm{K})$} & 35.32                             & 17.41                             & 32.79                             \\
                               & \textbf{$\bm{O} = \mathrm{Attn}(\bm{Q}, \bm{K}, \bm{V})$} & 141.28                            & 89.81                             & 112.65                            \\
                               & \textbf{$\bm{X} = \bm{O} \bm{W}_O$}                       & 81.13                             & 38.52                             & 25.75                             \\
                               & \textbf{$\bm{X} = \mathrm{Add}\&\mathrm{Norm}(\bm{X})$}   & 28.13                             & 17.59                             & 18.47                             \\
                               & \textbf{$\bm{G}, \bm{U} = \bm{X} [\bm{W}_G, \bm{W}_U]$}   & 657.61                            & 275.45                            & 175.76                            \\
                               & \textbf{$\bm{D} = \mathrm{Swish}(\bm{G}) \odot \bm{U}$}    & 17.07                             & 12.53                             & 9.23                              \\
                               & \textbf{$\bm{X} = \bm{D} \bm{W}_D$}                       & 214.90                            & 91.07                             & 55.85                             \\
                               & \textbf{$\bm{X} = \mathrm{Add}\&\mathrm{Norm}(\bm{X})$}   & 28.13                             & 17.59                             & 18.47                             \\ \midrule
\multirow{9}{*}{\textbf{vLLM}} & \textbf{$\bm{Q}, \bm{K}, \bm{V} = \bm{X} \bm{W}_{QKV}$}   & 252.45                            & 100.47                            & 84.92                             \\
                               & \textbf{$\bm{Q}, \bm{K} = \mathrm{RoPE}(\bm{Q}, \bm{K})$} & 6.35                              & 4.03                              & 6.23                              \\
                               & \textbf{$\bm{O} = \mathrm{Attn}(\bm{Q}, \bm{K}, \bm{V})$} & 33.26                             & 20.10                             & 22.74                             \\
                               & \textbf{$\bm{X} = \bm{O} \bm{W}_O$}                       & 87.48                             & 36.34                             & 22.95                             \\
                               & \textbf{$\bm{X} = \mathrm{Add}\&\mathrm{Norm}(\bm{X})$}   & 6.74                              & 3.86                              & 4.87                              \\
                               & \textbf{$\bm{G}, \bm{U} = \bm{X} [\bm{W}_G, \bm{W}_U]$}   & 453.50                            & 184.23                            & 124.33                            \\
                               & \textbf{$\bm{D} = \mathrm{Swish}(\bm{G}) \odot \bm{U}$}    & 12.28                             & 8.49                              & 8.92                              \\
                               & \textbf{$\bm{X} = \bm{D} \bm{W}_D$}                       & 231.07                            & 98.01                             & 63.29                             \\
                               & \textbf{$\bm{X} = \mathrm{Add}\&\mathrm{Norm}(\bm{X})$}   & 6.74                              & 3.86                              & 4.87                              \\ \bottomrule
\end{tabular}
\caption{
Detailed running time (ms) of Transformers and vLLM when varying hardware in the prefill stage ($b = 8, s = 512$).
}
\label{exp-prefill-time-gpu}
\end{table}

\begin{table}[H]
\centering
\begin{tabular}{llrrr}
\toprule
                                &                                                            & \multicolumn{1}{c}{\textbf{3090}} & \multicolumn{1}{c}{\textbf{4090}} & \multicolumn{1}{c}{\textbf{A800}} \\ \midrule
\multirow{10}{*}{\textbf{TRF}}  & \textbf{$\bm{q}, \bm{k}, \bm{v} = \bm{x} \bm{W}_{QKV}$}    & 5.28                              & 3.86                              & 2.72                              \\
                                & \textbf{$\bm{q}, \bm{k} = \mathrm{RoPE}(\bm{q}, \bm{k})$}  & 1.79                              & 1.04                              & 2.66                              \\
                                & \textbf{$\bm{K}, \bm{V} = \mathrm{Cache}(\bm{k}, \bm{v})$} & 11.16                             & 6.41                              & 10.89                             \\
                                & \textbf{$\bm{o} = \mathrm{Attn}(\bm{q}, \bm{K}, \bm{V})$}  & 3.64                              & 3.66                              & 3.52                              \\
                                & \textbf{$\bm{x} = \bm{o} \bm{W}_O$}                        & 1.76                              & 1.29                              & 0.91                              \\
                                & \textbf{$\bm{x} = \mathrm{Add}\&\mathrm{Norm}(\bm{x})$}    & 1.26                              & 0.96                              & 1.83                              \\
                                & \textbf{$\bm{g}, \bm{u} = \bm{x} [\bm{W}_G, \bm{W}_U]$}    & 7.13                              & 6.35                              & 3.87                              \\
                                & \textbf{$\bm{d} = \mathrm{Swish}(\bm{g}) \odot \bm{u}$}          & 0.20                              & 0.16                              & 0.27                              \\
                                & \textbf{$\bm{x} = \bm{d} \bm{W}_D$}                        & 4.35                              & 3.23                              & 2.05                              \\
                                & \textbf{$\bm{x} = \mathrm{Add}\&\mathrm{Norm}(\bm{x})$}    & 1.26                              & 0.96                              & 1.83                              \\ \midrule
\multirow{10}{*}{\textbf{vLLM}} & \textbf{$\bm{q}, \bm{k}, \bm{v} = \bm{x} \bm{W}_{QKV}$}    & 3.90                              & 3.59                              & 2.11                              \\
                                & \textbf{$\bm{q}, \bm{k} = \mathrm{RoPE}(\bm{q}, \bm{k})$}  & 0.22                              & 0.17                              & 0.31                              \\
                                & \textbf{$\bm{K}, \bm{V} = \mathrm{Cache}(\bm{k}, \bm{v})$} & 0.26                              & 0.19                              & 0.38                              \\
                                & \textbf{$\bm{o} = \mathrm{Attn}(\bm{q}, \bm{K}, \bm{V})$}  & 2.84                              & 2.57                              & 1.81                              \\
                                & \textbf{$\bm{x} = \bm{o} \bm{W}_O$}                        & 1.75                              & 1.26                              & 0.89                              \\
                                & \textbf{$\bm{x} = \mathrm{Add}\&\mathrm{Norm}(\bm{X})$}    & 0.20                              & 0.15                              & 0.26                              \\
                                & \textbf{$\bm{g}, \bm{u} = \bm{x} [\bm{W}_G, \bm{W}_U]$}    & 8.06                              & 6.35                              & 3.68                              \\
                                & \textbf{$\bm{d} = \mathrm{Swish}(\bm{g}) \odot \bm{u}$}          & 0.27                              & 0.20                              & 0.42                              \\
                                & \textbf{$\bm{x} = \bm{d} \bm{W}_D$}                        & 4.33                              & 3.17                              & 2.02                              \\
                                & \textbf{$\bm{x} = \mathrm{Add}\&\mathrm{Norm}(\bm{x})$}    & 0.20                              & 0.15                              & 0.26                              \\ \bottomrule
\end{tabular}
\caption{
Detailed running time (ms) of Transformers and vLLM when varying hardware in the decoding stage ($b = 8, s = 512$).
}
\label{exp-decoding-time-gpu}
\end{table}


\begin{table}[H]
\centering
\small
\begin{tabular}{lcccccc}
\toprule
\textbf{Libs.} & \textbf{$\alpha$}      & \textbf{$\beta$}       & \textbf{$\gamma$}      & \textbf{$\eta$}       & \textbf{$\lambda$}    & \textbf{$\mu$}      \\ \midrule
\textbf{TRF}   & $3.75 \times 10^{-11}$ & $3.69 \times 10^{-11}$ & $4.20 \times 10^{-8}$  & $1.70 \times 10^{-7}$ & $6.35 \times 10^{-9}$ & $3.28 \times 10^{1}$  \\
\textbf{vLLM}  & $4.51 \times 10^{-11}$ & $3.35 \times 10^{-11}$ & $2.29 \times 10^{-9}$  & $5.88 \times 10^{-8}$ & $6.26 \times 10^{-9}$ & $-1.64 \times 10^{0}$ \\ \midrule
\textbf{Libs.} & \textbf{$\phi$}        & \textbf{$\psi$}        & \textbf{$\omega$}      & \textbf{$\nu$}      &                       &                       \\ \midrule
\textbf{TRF}   & $2.31 \times 10^{-8}$  & $2.65 \times 10^{-11}$ & $3.32 \times 10^{-12}$ & $1.85 \times 10^{1}$  &                       &                       \\
\textbf{vLLM}  & $2.23 \times 10^{-9}$  & $1.75 \times 10^{-11}$ & $1.63 \times 10^{-8}$  & $1.12 \times 10^{1}$  &                       &                       \\ \bottomrule
\end{tabular}
\caption{
The coefficients of running time (ms) estimation Equation~\ref{eq:time-formula}.
}
\label{exp-time-formula}
\end{table}